\documentclass[lettersize,journal]{IEEEtran}
\usepackage{amsmath,amsfonts}
\usepackage{algorithmic}
\usepackage{algorithm}
\usepackage{amsmath}
\usepackage{array}
\usepackage[caption=false,font=normalsize,labelfont=sf,textfont=sf]{subfig}
\usepackage{textcomp}
\usepackage{stfloats}
\usepackage{url}
\usepackage{verbatim}
\usepackage{graphicx}
\usepackage{bm}
\usepackage{cite}
\usepackage{hyperref}
\newcommand{\INPUT}{\item[\textbf{Input:}]}  
\newcommand{\OUTPUT}{\item[\textbf{Output:}]}  
\usepackage{array}
\usepackage{booktabs}
\begin{document}

\title{Semantic Successive Refinement: A Generative AI-aided Semantic Communication Framework}
\author{
	\IEEEauthorblockN{
		Kexin Zhang\IEEEauthorrefmark{1}, 
		Lixin Li\IEEEauthorrefmark{1}, 
		Wensheng Lin\IEEEauthorrefmark{1}, 
		Yuna Yan\IEEEauthorrefmark{1}, 
		Rui Li\IEEEauthorrefmark{2}, 
		Wenchi Cheng\IEEEauthorrefmark{3}, 
		and Zhu Han\IEEEauthorrefmark{4}
	} 
	
	\IEEEauthorblockA{\IEEEauthorrefmark{1}School of Electronics and Information, Northwestern Polytechnical University, Xi’an, China, 710129}
	
	\IEEEauthorblockA{\IEEEauthorrefmark{2}Samsung AI Cambridge Center, Cambridge CB1 2RE, U.K.}
	
	\IEEEauthorblockA{\IEEEauthorrefmark{3}State Key Laboratory of Integrated Services Networks, Xidian University, Xi’an, China, 710071}
	
	\IEEEauthorblockA{\IEEEauthorrefmark{4}Department of Electrical and Computer Engineering, University of Houston, Houston, TX, 77004} 
}


\maketitle
	
\begin{abstract}
Semantic Communication (SC) is an emerging technology aiming to surpass the Shannon limit. 
Traditional SC strategies often minimize signal distortion between the original and reconstructed data, neglecting perceptual quality, especially in low Signal-to-Noise Ratio (SNR) environments. 
To address this issue, we introduce a novel Generative AI Semantic Communication (GSC) system for single-user scenarios. This system leverages deep generative models to establish a new paradigm in SC. Specifically, At the transmitter end, it employs a joint source-channel coding mechanism based on the Swin Transformer for efficient semantic feature extraction and compression. At the receiver end, an advanced Diffusion Model (DM) reconstructs high-quality images from degraded signals, enhancing perceptual details.
Additionally, we present a Multi-User Generative Semantic Communication (MU-GSC) system utilizing an asynchronous processing model. This model effectively manages multiple user requests and optimally utilizes system resources for parallel processing. Simulation results on public datasets demonstrate that our generative AI semantic communication systems achieve superior transmission efficiency and enhanced communication content quality across various channel conditions. Compared to CNN-based DeepJSCC, our methods improve the Peak Signal-to-Noise Ratio (PSNR) by 17.75\% in Additive White Gaussian Noise (AWGN) channels and by 20.86\% in Rayleigh channels.

\end{abstract}


\begin{figure*}[!t]
	\centering
	\includegraphics[width=\textwidth]{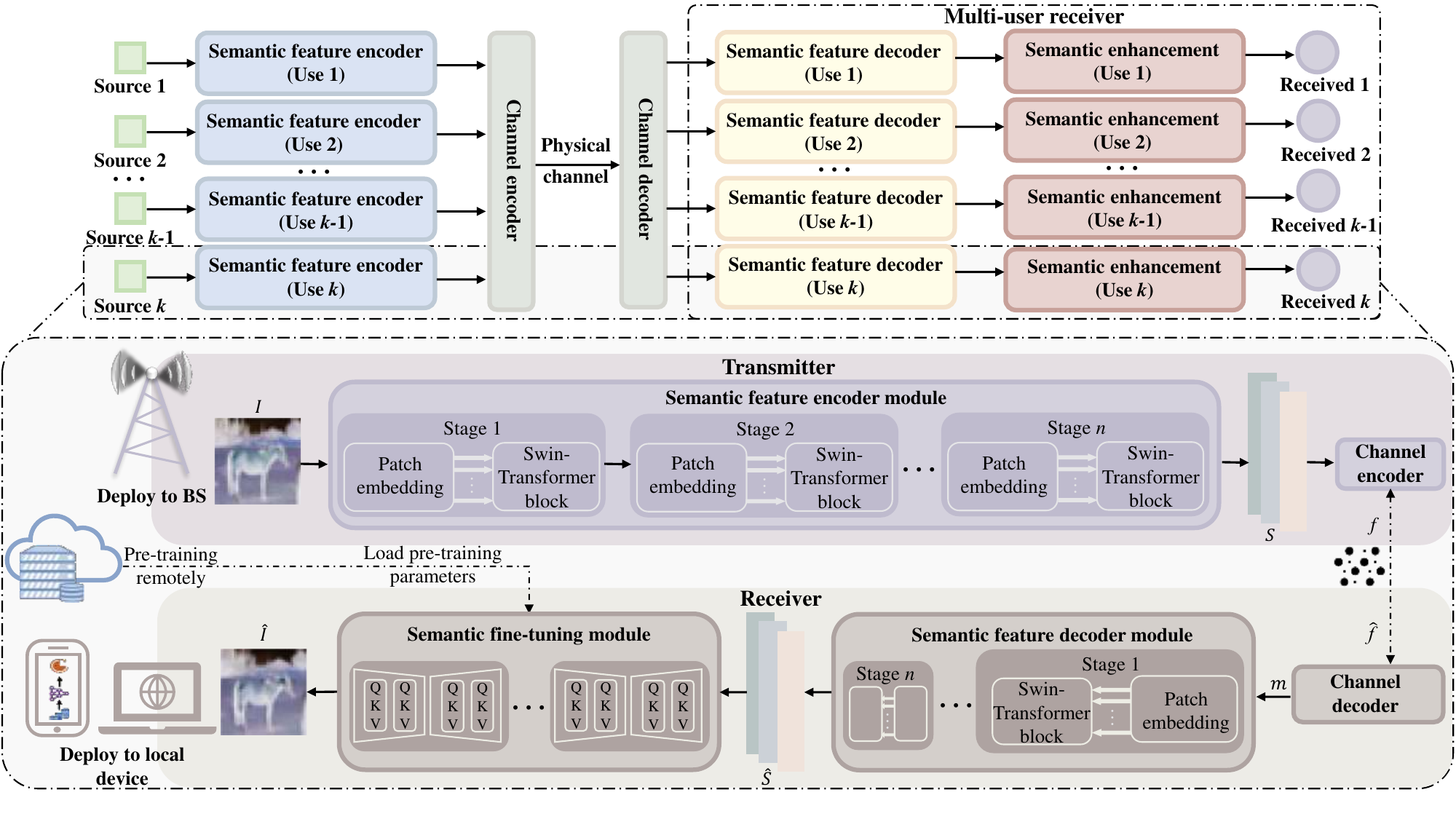}
	\caption{The overview of the proposed generative AI semantic communication system.}
	\label{fig1}
\end{figure*}

\section{Introduction}
\IEEEPARstart{W}{ith} the rapid growth of wireless communication technology, data traffic and mobile device connectivity is increasing exponentially. Semantic Communications (SC) is playing a pivotal role in this progress \cite{ref2}. Unlike traditional communication methods that focus on transmitting binary bits, SC emphasizes the context of the information \cite{add3}. Recent advancements further highlight SC's potential. For instance, the semantic interference cancellation (SemantIC) technique enhances information quality by iteratively eliminating noise in both the signal and semantic domains without additional channel resource costs \cite{add4lin}. Similarly, the Semantic-Forward (SF) relaying framework improves network robustness and reduces forwarding payload by extracting and transmitting semantic features, even under adverse channel conditions \cite{add5lin}.

Simultaneously, the rapid advancement of generative Artificial Intelligence (AI) models and the widespread adoption of technologies such as ChatGPT, Imagen, Midjourney, and DALL-E have prompted the B5G and 6G communities to synergize with these cutting-edge generative AI technologies. This paradigm shift is particularly beneficial for the vast amount of multimedia content generated by these applications, such as images and videos. Typically, AI models run on powerful cloud servers due to their high computational demands. However, with the growing prevalence of mobile devices, AI companies strive to provide high-quality AI-Generated Content (AIGC) services accessible from anywhere \cite{ref3, addLRui}. These innovations lay a solid foundation for further integrating generative AI with SC, which can significantly boost network performance. 

By considering the context of the generated content, SC can effectively leverage these advanced generative AI models to enhance overall system performance.
For example, consider a user playing an online game on their VR/AR device at home, entering a new virtual scene rendered by a Diffusion Model (DM) in the cloud. The scene is represented by information bits communicated to the VR/AR glasses via mobile networks. A generative AI-aware semantic communication network can utilize the fact that DM generates this content and transmits the latent representation of the content while performing the stable diffusion process on the receiver’s end. In this scenario, the semantic communication network focuses on conveying the meaning (latent representations) of the generative AI content rather than optimizing solely the delivery of the 0s and 1s. This approach allows for quality service metrics that align more closely with human visual perception than traditional bit-wise metrics. 

However, directly combining semantic communication with generative AI in fixed structures is inefficient due to the inability to adjust semantic density. In this context, deploying DM in the decoder offers a promising solution to accurately recover the image context.
Toward this end, we exploit the potential of the generative model and propose a novel generative AI semantic communication system, which aims to incorporate the robust, stable diffusion algorithm into the semantic decoding process as an initial step towards a fully collaborative generative AI semantic communication system. The main contributions of this paper are as follows:

\begin{itemize}
	
	\item To significantly enhance the efficiency and reliability of semantic communication systems in the era of generative AI, we propose a single-user Generative AI Semantic Communication (GSC) system powered by AI-generated content. Specifically, the Base Station (BS) encodes images using a joint source-channel coder based on the Swin Transformer. A diffusion model network is integrated at the receiver end to facilitate rich semantic reconstruction during wireless image transmission.  
	
	\item Existing research on semantic communication has predominantly focused on single-user scenarios, limiting its applicability in more complex, real-world environments. To address the challenges of scaling semantic communication to multi-user scenarios, we propose a Multi-User Generative Semantic Communication system (MU-GSC) that incorporates asynchronous concurrent processing, task parallel processing, and caching mechanisms to optimize multi-user communication efficiency.
	
	\item Classical diffusion models excel in image synthesis, but they require numerous iterative steps to estimate feature mappings, making them inefficient for direct use in communication systems. To address this concern, we use the strong distribution mapping abilities of diffusion models to guide semantic recovery at the decoding end by estimating a compact conditional vector. This approach significantly advances the connection between semantic communication and cutting-edge generative models, improving the end-to-end perceptual performance of wireless communication systems.
	
	\item Simulation experiments were conducted on public datasets. The results demonstrate that our proposed generative AI semantic communication system surpasses the benchmark communication system in data integrity and communication reliability. Additionally, the simulation results confirm that our designed multi-user communication system efficiently handles requests from multiple users, showcasing excellent performance and scalability.
	
\end{itemize}

This paper is structured as follows. Section \ref{sec2} provides a concise overview of related literature. Section \ref{sec3} provides a more detailed explanation of our system model. Following that, Section \ref{sec4} presents the simulation results and analysis of the proposed model. Finally, section \ref{sec5} summarizes the paper.

\section{Related Works}\label{sec2}

Transmitting image semantic information requires more communication resources than text data. 
Consequently, this section focuses on Image Semantic Communication (ISC). 
Research in this field can be divided into two main directions: semantic-oriented and task-oriented communication. 
The challenge of semantic-oriented communication lies in extracting and recovering semantics before and after transmission. 
In contrast, task-oriented communication focuses on completing specific tasks rather than accurately recovering all semantic information. 
The relevant works for these paradigms are outlined below.

\subsection{Image semantic communication}
Deep Joint Source-Channel Coding (DeepJSCC) \cite{ref4} is a pioneering study in wireless image transmission. This technique simplifies the code design process by integrating source and channel coding into a single mapping and eliminating the need for constellation diagrams used in digital schemes. Building on this foundation, the studies in \cite{ref5,ref6,ref7} extended DeepJSCC to various channel conditions. Yang \textit{et al.} \cite{ref6} proposed an adaptive wireless image transmission scheme that dynamically adjusts the transmission rate according to the current channel state and image complexity. Xu \textit{et al.} \cite{ref7} introduced a joint source-channel coding method incorporating an attention mechanism (ADJSCC). This framework adapts to noise by dynamically adjusting the number of bits allocated to the channel and source coder to maintain transmission reliability in real-world environments. However, these approaches primarily focus on the distortion of the reconstructed signal at the receiver relative to the source at the transmitter without adequately considering the perceptual quality of the reconstructed image, which may lead to significant perceptual distortion under extreme conditions such as low bandwidth and low Signal-to-Noise Ratio (SNR).

To address this, Kurka \textit{et al.} \cite{ref8} proposed DeepJSCC-f, which employs a feedback mechanism that allows the system to acquire real-time channel noise information and mitigate its effects by feeding the received signal back to the transmitter. However, this method assumes that any complex value can be transmitted over the channel, which may hinder the application of these algorithms in scenarios where the hardware or protocol only accepts certain sets of channel inputs (e.g., digital constellations). Additionally, the semantic communication systems proposed in \cite{ref9,ref10,ref11,ref12} are designed for receivers with powerful computational capabilities, enabling large-scale deep learning networks and single-user communication scenarios. 
Among them, Zhang \textit{et al.} \cite{ref9} developed a system that adaptively identifies and transmits task-specific key semantic features in a changing environment. Yang \textit{et al.} \cite{ref10} proposed the WITT framework to enhance CNN performance by introducing a spatial modulation module, which adjusts the scale of potential representation based on channel state information, improving the model's adaptability to different channel conditions. Yu \textit{et al.} \cite{ref11} extended the semantic communication system to bi-directional communication for IoT devices with limited capacity, significantly reducing training overheads by eliminating the need for information feedback and model migration.
Nguyen \textit{et al.} \cite{ref12} proposed a system that adapts to different computational capabilities by dynamically adjusting the transmission length of the output from the channel coder, combined with hybrid loss optimization, to achieve high-quality image reconstruction and avoid network congestion.

Task-oriented approaches optimize resource utilization by delivering information according to specific task requirements. Kadam \textit{et al.} \cite{ref13} designed a keyword-based SC system for transmitting and sharing knowledge recovery data for a specific Data Allocation Problem (DAP) to enhance efficiency and accuracy. Xie \textit{et al.} \cite{ref14} developed a multi-user system supporting two unimodal tasks (image retrieval and machine translation) and one multimodal task (a visual quiz combining text and images). Kang \textit{et al.} \cite{ref15} created a framework that enables users to retrieve image-related semantic information via text queries. However, this approach fails to fully consider the dynamic changes in the importance of semantic information, which can lead to information loss in resource competition. In \cite{ref16}, researchers employed a course-learning strategy to minimize the length of transmitted messages, aiming to improve communication efficiency for specific tasks. 

To address the issue of limited resource blocks from BS to individual users, Wang \textit{et al.} \cite{ref17} utilized a reinforcement learning algorithm based on an attention mechanism to prioritize the transmission of the most informative triples during resource allocation. Thomas \textit{et al.} \cite{ref18} emphasized the reasoning capabilities required by both the transmitter and receiver, employing a Generative Flow Network (GFlowNet) to achieve causal reasoning for bursty semantic communication with minimal data. Yoo \textit{et al.} \cite{ref19} demonstrated the feasibility of ISC in real-time wireless communication using Field Programmable Gate Array (FPGA) as a hardware platform, comparing its performance with conventional 256 Quadrature Amplitude Modulation (QAM) and showing superior application potential. 
However, their results were only validated in a simulation environment. 

\subsection{Generative AI semantic communication}
Generative AI revolutionizes visual computing by allowing users to programmatically create or modify realistic, high-quality images, videos, and 3D models. Neural network-driven generative models, such as Variational Autoencoders (VAEs), Generative Adversarial Networks (GANs), and Diffusion Models (DMs), have shown exceptional capabilities in generating high-quality data. This highlights their significant potential for application in joint source-channel coding.

Choi \textit{et al.} \cite{ref20} was the first to implement a Joint Source-Channel Coding (JSCC) scheme over binary channels using a discrete VAE. 
Subsequently, Hu \textit{et al.} \cite{ref21} employed an adversarial training approach and introduced the Masked Variational Quantized Variational Autoencoder (MVQ-VAE) to enhance the system's robustness to various noise types. 
Paper \cite{ref22} developed an innovative architecture for DeepJSCC that is data-driven and enhanced by adversarial training, with the combined goal of maximizing the reconstruction quality for legitimate receivers while minimizing adversarial losses. 
Yang \textit{et al.} \cite{ref23} combined an autoencoder with Orthogonal Frequency Division Multiplexing (OFDM), using a GAN-inspired loss function to efficiently train a robust decoder against the effects of multipath fading. 
The approach described in \cite{ref24} significantly reduces bandwidth requirements by performing semantic segmentation at the transmitter's end to extract semantic information from an image and using GAN for image reconstruction at the receiver's end. 
However, this reconstruction process may introduce subtle differences between the generated image and the original scene. 
To this end, Erdemir \textit{et al.} \cite{ref25} proposed two schemes: InverseJSCC, which addresses the inverse problem of DeepJSCC, and GenerativeJSCC, an end-to-end optimization scheme based on GANs that can reconstruct perceptual quality under extremely adverse channel conditions like never before.

Recent advancements in DMs have established new benchmarks in density estimation and sample quality, surpassing other generative models. Parametric Markov chains enhance the variational lower bound of the likelihood function, thereby enabling samples to represent the target distribution \cite{ref26} more accurately. DMs iteratively refine these samples through a methodical denoising process until the desired output is achieved.
Niu \textit{et al.} \cite{ref27} introduced a strategy based on DeepJSCC, their method initially adds independent Gaussian noise to the input image and then applies a diffusion process to introduce additional information components during decoding. 
In contrast, other studies \cite{ref28, ref29} utilize the denoising capabilities of DMs to help the receiver mitigate noise interference in the channel or semantic vectors. Furthermore, \cite{ref30} addresses the communication problem as an inverse problem and proposes a generative AI semantic communication framework that resolves the denoising or colouring issues in the low-dimensional latent representations of samples.
Chen \textit{et al.} \cite{ref31} proposed CommIN, an innovative framework that combines an Invertible Neural Network (INN) with a DM to model channel characteristics and degradation phenomena introduced by DeepJSCC. 
Grassucci \textit{et al.} \cite{ref32} developed a generative AI semantic communication framework that enhances the quality of inferred images by incorporating semantic chunks for rapid denoising. 
However, DMs still face practical challenges, such as large network size and the iterative nature of the process, which complicates training with limited computational resources. This study addresses these issues by deploying a lightweight DM on the receiver of a semantic communication system. By estimating compact conditional vectors instead of all pixels, the proposed method leverages the excellent mapping capabilities of DMs to achieve high-quality image reconstruction.

\begin{figure*}[!t]
	\centering
	\includegraphics[width=\textwidth]{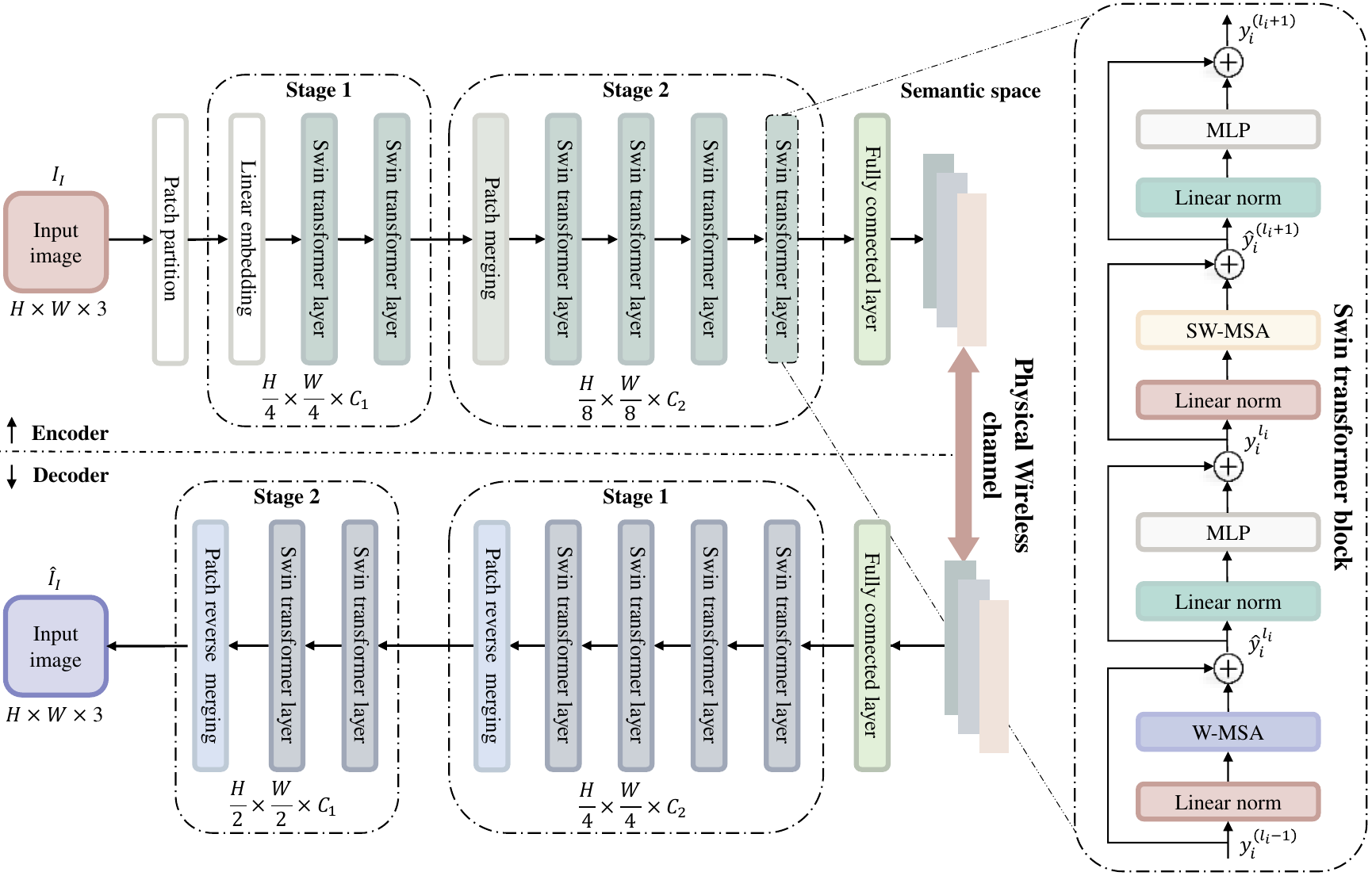}
	\caption{The architectures of the semantic communication network based on Swin Transformer.}
	\label{fig2}
\end{figure*}

\section{Proposed Method}\label{sec3}
Considering downlink transmission scenarios, this section describes the proposed generative AI semantic communication model. 
Fig. \ref{fig1} illustrates the general framework of the proposed approach. The architecture of multi-user generative communication system MU-GSC is fundamentally similar to that of single-user generative AI semantic communication system, which comprises four main processes:{\em the semantic feature encoder module, the physical wireless channel module, the semantic feature decoder module, and the semantic fine-tuning module.} 
Building on the foundation of GSC, the development strategy for MU-GSC involves creating a system based on an asynchronous processing model. This system is designed to handle requests from multiple users, optimally utilizing system resources for efficient parallel processing. The details for the implementation are presented.

\subsection{Semantic feature encoder module}
The BS usually consists of two modules in semantic communication systems: a semantic feature encoder and a channel coder. The semantic feature encoder, also known as the source encoder, mines information and extracts features from the images based on the knowledge base. Let $I$ be the transmitted image sources and $S$ be the extracted semantic symbols. The mathematical description is as follows:
\begin{equation}
	\label{deqn_ex1a}
	S = \boldsymbol{E}(I; \varphi_{\alpha}), I \in \mathbb{R}^{n},
\end{equation}
\noindent where $\boldsymbol{E}\left(  \cdot  \right)$ denotes the semantic coder network, $\varphi_{\alpha}$ is the parameter set of the corresponding coding network, and $n$ is the dimension of the input images.

In a semantic communication system, the edge server is responsible for training the local model, and the training process requires the encoder and decoder to be coupled alternately and in an end-to-end manner. Here, the semantic encoder is first modeled, and the semantic decoder follows the reverse architecture of the encoder.

\begin{figure*}[!t]
	\centering
	\includegraphics[width=\textwidth]{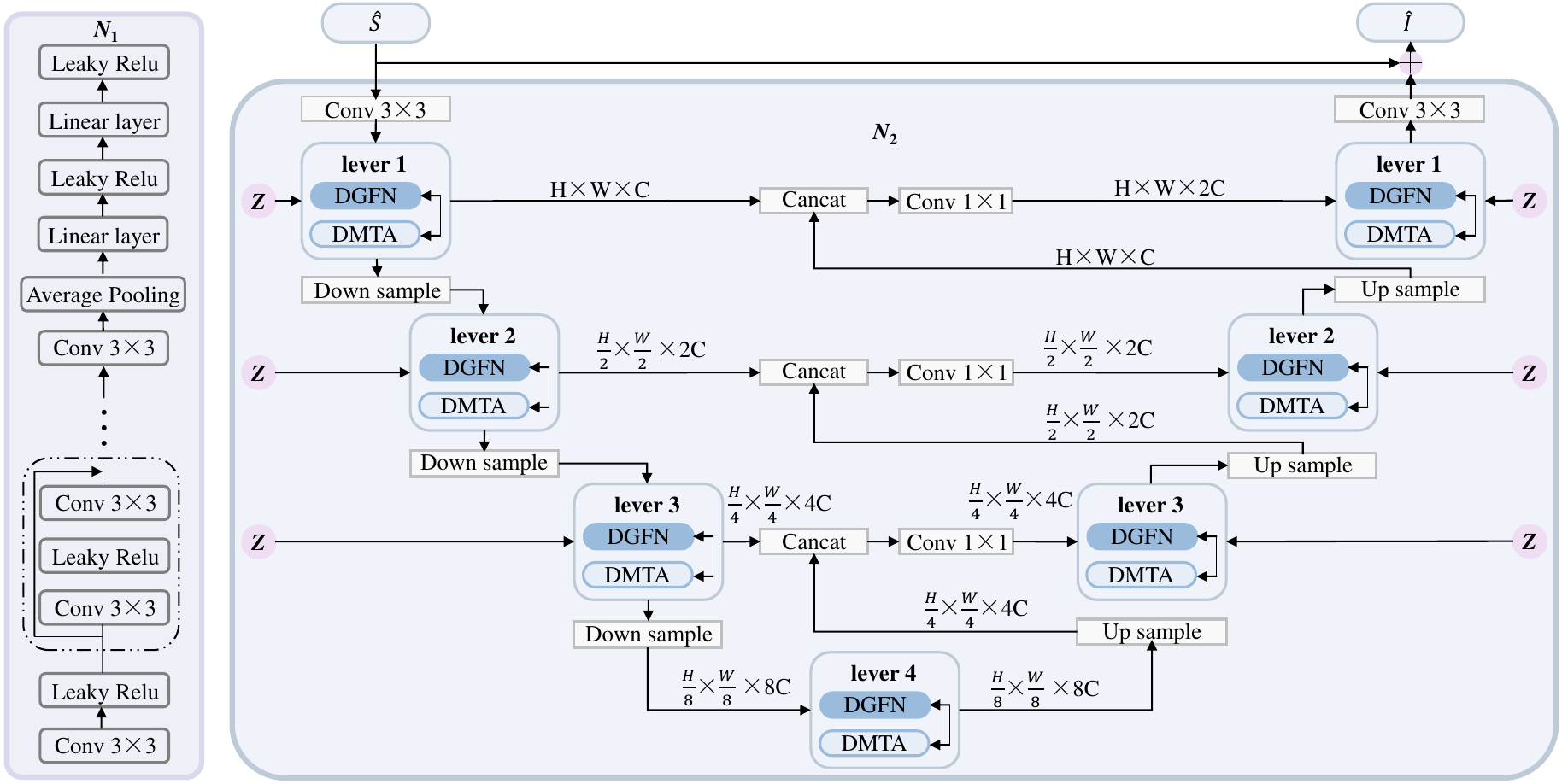}
	\caption{The architectures of the prior extraction network ${\bm{N}_1}$ and the image denoising network ${\bm{N}_2}$.}
	\label{figN}
\end{figure*}

The codec network in this paper is constructed based on the state-of-the-art visual Swin Transformer \cite{ref33}. On the transmitter, given a set of image sets $\boldsymbol{I} = \left\{ I_i \right\}_{i = 1}^N$, where $I_i \in \mathbb{R}^{3 \times H \times W}$ denotes the \(i\)-th image, $N$ denotes the size of the image set, $H$ and $W$ denote the height and width of the images, respectively.
The patch size is $4 \times 4$, which is partitioned into non-overlapping patches $s = \frac{H}{4} \times \frac{W}{4}$ by the patch partition module. Then, we perform a linear embedding by arranging these tokens in a sequence $\left( {{I_1},{I_2}, \cdots {I_s}} \right)$ from the top left to the bottom right, which projects the tokens into an embedding representation $\left( {\frac{H}{4},\frac{W}{4},C} \right)$ of arbitrary dimension $C$. After the patch embedding, the $s$ tokens are then input into the Swin Transformer block. This process is called ``stage 1".

From Fig. \ref{fig2}, it can be seen that the first patch merging layer and the Swin Transformer block are merged into ``stage 2". 
The patch merge layer splices each group of neighboring patches of size $2 \times 2$ so that the number of patch tokens becomes $1/4$ of the original one, i.e. $\frac{H}{8} \times \frac{W}{8}$. At the same time, the dimension of the patch token is expanded by a factor of $4$, i.e.$4C$. In order to reduce the output dimension and realize the downsampling of the feature map, the patch merging layer then performs the fully connected operation. after this process, the dimension of the concatenated feature patch is reduced $2C$ from $4C$. Then, the patch is feature transformed by Swin Transformer block. Finally, the output dimension becomes $\left( {\frac{H}{8},\frac{W}{8},C} \right)$. Considering the computational power of the hardware and the uncertainty of the input graph, here the semantic decoder contains a two-stage Swin Transformer architecture, and in general, high-resolution images require more stages.

The Swin Transformer module is a sequence-to-sequence function that consists of multiple Swin Transformer layers space. The right side of Fig. \ref{fig2} shows the structure of the Swin Transformer layers.Each layer consists of a window polytope layer and a moving window polytope layer.Since the two sub-layers have the same dimensions of inputs and outputs, they are sequentially connected of the two sub-layers. The use of the shifted-window partitioning approach can significantly enhance the modeling capability. The successive Swin Transformer blocks ${l_i}$ layer and ${l_i} + 1$ layer for ``stage $i$" are computed as:
\begin{equation}
	\label{deqn_ex2a}
	\begin{array}{l}
		\hat y_i^{{l_i}} = {\mathop{\rm W}\nolimits}  - MSA\left( {{\mathop{\rm LN}\nolimits} \left( {y_i^{{l_i} - 1}} \right)} \right) + y_i^{{l_i} - 1},\\
		{\rm{     }}y_i^{{l_i}} = {\mathop{\rm MLP}\nolimits} \left( {{\mathop{\rm LN}\nolimits} \left( {\hat y_i^{{l_i}}} \right)} \right) + \hat y_i^{{l_i}},\\
		\hat y_i^{{l_i} + 1} = {\mathop{\rm SW}\nolimits}  - MSA\left( {{\mathop{\rm LN}\nolimits} \left( {y_i^{{l_i}}} \right)} \right) + y_i^{{l_i}},\\
		{\rm{  }}y_i^{{l_i} + 1} = {\mathop{\rm MLP}\nolimits} \left( {{\mathop{\rm LN}\nolimits} \left( {\hat y_i^{{l_i} + 1}} \right)} \right) + \hat y_i^{{l_i} + 1},
	\end{array}
\end{equation}
\noindent Here, the MLP has two layers, the Window-based Multi-head Self-Attention (W-MSA) and the Shifted Windows Multi-head Self-Attention (SW-MSA) are multi-head self-concerned modules with regular and shifted window configurations.LN denotes the layer normalization operation. $\hat y_i^{{l_i}}$ and $y_i^{{l_i}}$ represent the output characteristics of the (S)W-MSA module and MLP module of module ${l_i}$ at ``stage $i$", respectively.

In order to protect the transmitted symbols from channel noise and interference in the wireless environment, the output will $S$ be further converted in the channel coder into a complex bit stream (vector) suitable for transmission over the physical channel. Here a non-trainable fully connected layer is used for simulation:
\begin{equation}
	\label{deqn_ex3a}
	f = \boldsymbol{C}(S; \varphi_\beta) = \boldsymbol{W}_n S + b_n, f \in \mathbb{R}^k,
\end{equation}
\noindent where $\boldsymbol{C}\left(  \cdot  \right)$ is the channel coder with parameter set $\varphi _\beta$ and $k$ is the length of $f$. The weight matrix $\boldsymbol{W}_n$ and the bias matrix $b_n$ determine the mapping relationship of the neural network. Here, $\boldsymbol{W}_n$ represents the channel gain, and $b_n$ simulates the noise in the channel by adding a random variable to the signal, the variance of the random variable represents the power of the channel noise, which is in turn constrained by the SNR and the transmit power.

This conversion process not only improves the signal's immunity to interference but also realizes an effective mapping from high-dimensional data to low-dimensional data, preserving the integrity and interpretability of the semantic content. The compression ratio is calculated by taking the ratio $k/n$ of the transmitted data size to the original image size. While it is possible to compress the semantic feature vectors further and thus reduce the transmission length, this simultaneously introduces the problem of the trade-off between the compressed size and the received image quality. Based on empirical values, this study selects 1/6 of the experimental conditions.

\subsection{Physical wireless channel module}
Considering the finite transmit power of the transmitting device, a power normalization operation must be used to make the signal $f$ satisfy the average power constraint before sending the transmitted signal to the channel. This implies $\frac{1}{k} \mathbb{E} {\rm{ }}{_f}\left[ {\left\| f \right\|_2^2} \right] \le P$.
Since the encoder and decoder are trained end-to-end, the physical channel can be modeled with a frozen neural network.
In this paper, we consider the general fading channel model with transfer function $\hat f = w{\rm{ }}(f;h) = h \odot f + n$, where $\odot$ is the element-wise product, $h$ denotes the Channel State Information (CSI) vector, and each component of the noise vector $n$ is independently sampled from a Gaussian distribution, i.e., $n\sim {\cal N}\left( {0,{\sigma ^2}I} \right)$, where ${\sigma ^2}$ is the average noise power.

\subsection{Semantic feature decoder module}
A joint source-channel decoder will be deployed at the receiver of the local device, including two modules, the channel decoder and the semantic decoder. The signal is first binary converted by the channel decoder and then sent to the semantic decoder to reduce from a potential representation of noise to a semantic feature map that the user cannot understand.

Channel distortion and noise are critical factors that cannot be avoided when transmitting coded feature vectors in a wireless channel environment. Taking an Additive White Gaussian Noise (AWGN) channel as an example, the received signal vector can be written as:
\begin{equation}
	\label{deqn_ex4a}
	{\hat f} = {f} + {\varepsilon},
\end{equation}
\noindent where ${\varepsilon}$ is a noise vector whose elements obey ${\cal N}\left( {0,{\sigma ^2}I} \right)$.
\begin{equation}
	\label{deqn_ex5a}
	m = \boldsymbol{C}^{-1}\left( \hat{f}; \varphi_{\phi} \right),
\end{equation}
\noindent where ${\hat f}$ is the estimated feature vector of the transmitted image source $I$, $\boldsymbol{C}^{ - 1}\left(  \cdot  \right)$ is the channel decoder, and $r$ is the data bits recovered from the code word $\hat f$ received from the channel.

The semantic decoder receives the signal disturbed. This process can be represented as:
\begin{equation}
	\label{deqn_ex6a}
	\hat S = {\boldsymbol{E}^{ - 1}}\left( {m;{\varphi _\gamma }} \right),
\end{equation}
\noindent where $\boldsymbol{E}^{ - 1}$ denotes the semantic decoder network, $\varphi _\gamma$ is the parameter set of the corresponding network, and $\hat S$ is the decoded semantic information images.

Most of the existing research on semantic image enhancement has used the MSE loss as the objective for the training process, which has the property of being continuously differentiable across any domain. Therefore, the MSE between the original input images and the user's received images is used as the loss function in the training phase, and $l$ is the length of the image vector, which is obtained from the product of the height, width and number of channels of the images:
\begin{equation}
	\label{deqn_ex7a}
	{{\cal L}_{dec}}\left( {I,{{\hat I}_k}} \right) = {\mathop{\rm MSE}\nolimits}  = \frac{1}{l}{\sum\limits_{i = 1}^l {\left( {{I_i} - {{\hat I}_i}} \right)} ^2}.
\end{equation}

\subsection{Semantic fine-tuning module}
Considering the significant increase in computing power of mobile devices, many user terminals can perform relatively simple fine-tuning operations. In this subsection, the Semantic Fine-Tuning (SFT) module is designed. Specifically, after the semantic features are decoded, they need to be further transferred to the DM with pre-training parameters to perform semantic enhancement. Inspired by image retrieval algorithms, adding precise details to low-quality images avoids the need for the DM to generate complete images. As a result, SFT achieves more accurate estimation with fewer iterations and ensures the stability of the results.

The training process of the SFT module primarily consists of pre-training and DM training, both conducted on the cloud server. Detailed information about this process is provided in Alg. \ref{alg1}. The pre-training phase involves two networks, as shown in Fig. \ref{figN}: the prior extraction network (${\bm{N}_1}$) and the image denoising network (${\bm{N}_2}$) \cite{ref26}. 
The primary function of ${\bm{N}_1}$ is to extract Prior Representations (PRs) in the form of conditional vectors. The workflow of the pre-training phase is as follows: we initially combine the transmitted images with the decoded images through concatenation. Following this, we utilize the PixelUnshuffle operation to downsample the concatenated images to serve as input for ${\bm{N}_1}$. The extracted PRs from this process are designated as $Z$. Then, ${\bm{N}_2}$ can use the extracted $Z$ to restore images, which is stacked with dynamic transformer blocks in the Unet shape. The dynamic transformer blocks consists of Dynamic Multi-head Transposed Attention (DMTA) and Dynamic Gated Feed-Forward Network (DGFN), which can use $Z$ as dynamic modulation parameters to add restoration details into feature maps, effectively aggregating both local and global spatial characteristics. The trained model obtained from this pre-training phase is denoted as $\mathcal{M}_1$. The image semantic enhancement is represented as follows: 
\begin{equation}
	\label{deqn_ex8a}
	\hat{I} = \boldsymbol{N_2}\left( Z; \varphi_{\varpi} \right), Z \in \mathbb{R}^{4C'},
\end{equation}
where ${{\varphi _\varpi }}$ are the parameters of $\boldsymbol{N_2}$.

Next, ${\bm{N}_1}$ and ${\bm{N}_2}$ are jointly optimized so that ${\bm{N}_2}$ can effectively use the PRs extracted by ${\bm{N}_1}$ to guide the image semantic enhancement. The loss function is defined as follows:
\begin{equation}
	\label{deqn_ex9a}
	{{\cal L}_{pre}} = {\left\| {I - \hat I} \right\|_1},
\end{equation}
where $I$ and ${\hat I}$ are the transmitted and received images, respectively. ${\left\|  \cdot  \right\|_1}$ is the ${L_1}$ paradigm.

In the DM training phase, the accurately recovered images are generated from the lossy decoded images mainly through the efficient data estimation function of diffusion model, and this process includes two critical aspects: the forward diffusion and the backward inference. First, the PRs of the decoded images, denoted as ${{Z_0}}$, are captured using the pre-trained ${\bm{N}_1}$, and the forward diffusion process of ${{Z_0}}$ is applied to the sample ${{Z_T}}$ through $T$ iterations. Each iteration is as follows:
\begin{equation}
	\label{eq:deqn_ex10a}
	q\left( Z_T \mid Z_0 \right) = \mathcal{N}\left( Z_T; \sqrt{\bar{\alpha}_T} Z_0, \left(1 - \bar{\alpha}_T\right)I \right),
\end{equation}
\noindent where ${\bar \alpha _t} = \prod\nolimits_{i = 1}^t {{\alpha _i}}$ is the cumulative product of ${\alpha _t}$, the scheduler gradually adds Gaussian noise at each time step $t \in \left[ {0,T} \right]$ until the semantic information of ${Z_0}$ becomes pure noise ${Z_0}$:
\begin{equation}
	\label{deqn_ex11a}
	{Z_t} = \sqrt {1 - {\beta _t}} {Z_{t - 1}} + \sqrt {{\beta _t}} \varepsilon,
\end{equation}
\noindent where $0 < {\beta _1} < {\beta _2} <  \cdots  < {\beta _T} < 1$ is a variance table with time-dependent constants and $\varepsilon  \sim {\cal N}\left( {0,\boldsymbol{O}} \right)$ is a Gaussian noise with unit matrix $\boldsymbol{O}$.

The forward diffusion process notifies the data. On the contrary, the backward inference is the denoising process. Traditional DMs start from pure noise, train Unet to learn the conditional probability distribution of real images, and gradually denoise until a generated image is obtained. However, traditional distributed denoising algorithms can only optimize the denoising network by randomly choosing a time step during the iteration process, which greatly increases the computational cost. In addition, to fully utilize the capability of ${\bm{N}_2}$, SFT performs all denoising iterations from a specific time step to obtain PRs and sends it to ${\bm{N}_2}$, which enables joint optimization with the denoising network ${\varepsilon _\theta }$. In joint optimization, it is also necessary to use ${\bm{N}_1}$ to obtain the PRs denoted as $D\in \mathbb{R}^{4C'}$ from the decoded images. Then, the noise at each time step $t$ is estimated using ${\varepsilon _\theta }$ and to obtain ${\hat Z_{t - 1}}$, i.e.,
\begin{equation}
	\label{deqn_ex12a}
	{\hat Z_{t - 1}} = \frac{1}{{\sqrt {{\alpha _t}} }}\left( {{{\hat Z}_t} - \frac{{1 - {\alpha _t}}}{{\sqrt {1 - {{\bar \alpha }_t}} }}{\varepsilon _\theta }\left( {{{\hat Z}_t},t} \right)} \right),
\end{equation}
where ${\varepsilon _\theta }$ is the prediction noise, ${\alpha _t} = 1 - {\beta _t}$, and ${\bar \sigma _t}$ is also a time-dependent constant for the step.

After $T$ iterations, \( \hat{Z} \in \mathbb{R}^{4C'} \) is generated. Since the trained model $\mathcal{M}_2$ only adds details for recovery, DM can obtain stable visual results after several iterations. After that, ${\bm{N}_2}$ utilizes \( \hat{Z} \) to recover the semantic information images \( \hat{S} \). 

The diffusion model aims to predict the noise distribution of the conditioned vectors of the decoded images with a loss function denoted as $\mathcal{L}_{\text{diff}} = \frac{1}{4C'} \sum_{i = 1}^{4C'} \left| \hat{Z}(i) - Z_0(i) \right|$. To achieve efficient generation and propagation of semantic information, ${\cal L} = {{\cal L}_{pre}} + {{\cal L}_{diff}}$ is used here for joint optimization.

After completing the training process, the testing process is conducted on local devices, as detailed in Algorithm \ref{alg2}. Notably, we do not input the transmitted images into either the ${\bm{N}_1}$ or the ${\bm{N}_2}$. Furthermore, only the reverse inference process of the DM is employed.

\begin{algorithm}[H]
	\caption{The Training Process on the Cloud Server.}\label{alg:alg1}
	\begin{algorithmic}[1]
		\INPUT The trained model $\mathcal{M}_1$, $\beta_t (t \in [1, T])$
		\OUTPUT The trained model $\mathcal{M}_2$
		\STATE Init: $\alpha_t = 1 - \beta_t$, $\bar{\alpha}_T = \prod_{i=0}^T \alpha_i$
		\FOR {$(\hat{S}, I)$}
		\STATE $Z=\text{${\bm{N}_1}$}(\text{PixelUnshuffle}(\text{Concat}(\hat{S}, I)))$
		\STATE \textbf{Forward Diffusion Process:}
		\STATE We sample $Z_T$ by\\
		$q\left( Z_T \mid Z_0 \right) = \mathcal{N}\left( Z_T; \sqrt{\bar{\alpha}_T} Z_0, \left(1 - \bar{\alpha}_T\right)I \right)$ 
		\STATE \textbf{Reverse Inference Process:}
		\STATE $\hat{Z_T} = Z_T$
		\STATE $D = \text{${\bm{N}_1}$}(\text{PixelUnshuffle}(\hat{S}))$
		\FOR {$t = T$ to $1$}
		\STATE $\hat{Z}_{t-1} = \frac{1}{\sqrt{\alpha_t}} \left( \hat{Z}_t - \frac{1 - \alpha_t}{\sqrt{1 - \bar{\alpha}_t}} \epsilon_\theta (\text{Concat}(\hat{Z}_t, t, D)) \right)$
		\ENDFOR
		\STATE $\hat{Z} = \hat{Z}_0$
		\STATE $\hat{I} = \text{${\bm{N}_2}$}(\hat{S}, \hat{Z})$
		\STATE Calculate $L_{\text{diff}}$ loss
		\ENDFOR
		\STATE Output the trained model $\mathcal{M}_2$
	\end{algorithmic}
	\label{alg1}
\end{algorithm}

\begin{algorithm}[H]
	\caption{The Testing Process on the Local Device.}\label{alg:alg2}
	\begin{algorithmic}[1] 
		\INPUT The trained model $\mathcal{M}_2$, decoded images \(\hat{S}\)
		\OUTPUT The received images \(\hat{I}\)
		\STATE \textbf{Init:} \(\alpha_t = 1 - \beta_t, \bar{\alpha}_T = \prod_{i=0}^T \alpha_i\)
		\STATE \textbf{Reverse Inference Process:}
		\STATE Sample \({Z}_T \sim \mathcal{N}(0,\boldsymbol{O})\)
		\STATE $D = \text{${\bm{N}_1}$}(\text{PixelUnshuffle}(\hat{S}))$
		\FOR{$t = T$ to $1$}
		\STATE \(\hat{Z}_{t-1} = \frac{1}{\sqrt{\alpha_t}} \left( \hat{Z}_t - \frac{1 - \alpha_t}{\sqrt{1 - \bar{\alpha}_t}} \epsilon_\theta \left( \text{Concat}(\hat{Z}_t, t, D) \right) \right) \quad\)		
		\ENDFOR
		\STATE \(\hat{Z} = \hat{Z}_0\)
		\STATE $\hat{I} = \text{${\bm{N}_2}$}(\hat{S}, \hat{Z})$
		\STATE Output received images \(\hat{I}\)
	\end{algorithmic}
	\label{alg2}
\end{algorithm}

\subsection{Multi-user communication system}
The multi-user scenario refers to each user transmitting independent semantic information to perform their transmission tasks. As shown in Fig. \ref{fig1}, a multi-user generative AI semantic communication system that consists of $k$ sources and $k$ destinations is considered. 

The message of the \(k\)-th user's source is denoted as $I_k$. Each source transmits semantic information. Similar to the single-user cognitive semantic communication system, the semantic information is first expressed as the semantic symbols. The semantic symbol $S_k$ is abstracted from the image source $I_k$ by using our proposed semantic feature encoder, given as:
\begin{equation}
	\label{deqn_ex13a}
	S_k = \boldsymbol{E}(I_k), k=1, 2, ... ,n.
\end{equation}

After the semantic symbols of each source are obtained, they are transmitted by exploiting the conventional non-trainable fully connected layer to simulate. Specifically, the semantic symbol $S_k$ is encoded in order to improve transmission efficiency, and $f_k$ is obtained, i.e.,
\begin{equation}
	\label{deqn_ex14a}
	f_k = \boldsymbol{C}(S_k) = \boldsymbol{W}_n S_k + b_n, k=1, 2, ... ,n,
\end{equation}
where $\boldsymbol{C}$ is the channel coding; $f_k$ and $S_k$ are the channel coding and semantic symbols of the \(k\)-th source, respectively.

After transmission over the channel, the channel decoding is performed at each user receiver, and the reconstructed semantic symbol ${\hat S_k}$ is obtained by exploiting our proposed semantic feature decoder module. Note that the semantic symbols of different users are not distinguished in this process. Thus, the reconstructed semantic symbol of each user is mixed, given as:
\begin{equation}
	\label{deqn_ex15a}
	\hat S_k = {\boldsymbol{E}^{ - 1}}\left( {\boldsymbol{C}^{-1}\left( \hat{f}_k\right)} \right), k=1, 2, ... ,n,
\end{equation}
where $\boldsymbol{C}^{ - 1}$ is the channel decoding and $\boldsymbol{E}^{ - 1}$ is the semantic feature decoding, $\hat S_k$ is the reconstructed semantic information images and $\hat f_k$ is the channel coding received vector at the \(k\)-th user.

Then, the reconstructed semantic symbols of each user need to be fine tuning. All models can be trained in the cloud and then broadcast to users. Similar to the single-user cognitive semantic communication system, the received images is obtained by exploiting our proposed semantic fine-tuning module.

To address the challenge of data processing in multi-user scenarios, we first implemented a data segmentation strategy during the preprocessing stage. This approach simulates different user sources generating diverse messages, enabling effective allocation of user tasks to various processing units and allowing these tasks to be executed concurrently. Subsequently, we introduced an asynchronous concurrent processing model. Employing asynchronous task processing functions and event loops ensured that the system's main thread remained unblocked during I/O operations, thereby facilitating the concurrent execution of multiple user tasks.

To further enhance system performance, we leveraged task parallel processing technology. This technique involves decomposing a single user task into smaller subtasks and executing these subtasks concurrently across multiple processing units. This approach fully utilizes GPU resources to boost system concurrency and performance. Additionally, to optimize system efficiency and reduce computational load, we incorporated a caching mechanism. This mechanism stores previously computed results and reuses them when needed, thus avoiding redundant calculations and enhancing the system's response speed and throughput. Through implementation, this multi-user communication system based on an asynchronous processing model significantly improves resource utilization, processing speed, and system stability. It is suitable for scenarios requiring parallel processing of a large number of user communication tasks, with a high practical application value.

\section{Simulation Results and Discussions}\label{sec4}
In this section, simulation results are presented to evaluate the performance of our proposed single-user and multi-user generative AI semantic communication systems. They are compared with deep learning-based methods and the traditional communication systems realized by the separate source and channel coding technologies. In addition, to demonstrate the robustness of the proposed method, the simulation experiments are performed under the AWGN channels and Rayleigh fading channels, where the perfect CSI is assumed for all methods. For the MU-GSC, the transceiver is assumed with three single-antenna users and the receiver with three antennas.

\subsection{Simulation setup}
The training process is divided into two independent parts: semantic feature encoder/decoder and SFT. To comply with the generalization of ISC, the classical communication network based on the Swin Transformer is chosen as the semantic feature coder in this paper. Specifically, this network uses the ReLU activation function between the input layer and the two hidden layers, the hyperparameter $C$ is set to $32$, the loss function adopts the MSE, the batch size is set to $32$, and the window size is set to $2$, and then it is trained using Adam's optimizer \cite{ref34} with a learning rate of $1 \times 10^{-4}$. 
During the SFT training process, we adopt a two-step training strategy utilizing a four-level encoder-decoder structure in the DTBN. From level $1$ to level $4$, the attention heads in DMTA are set to [$1, 2, 4, 8$], the number of dynamic transformer blocks to [$3, 5, 6, 6$] and the number of channels $C'$ of ${\bm{N}_1}$ is set to $96$. Training begins with a patch size of $32 \times 32$ and a batch size of 16. In the second step, \(\beta _t\) linearly increases from $0.10$ to $0.99$, starting with an initial learning rate of $2 \times 10^{-4}$, and the total timesteps $T$ are set at $4$.

In this simulation, the experimental platform for training and testing is built on an Ubuntu $20.04$ system with CUDA $11.8$ support, and the deep learning framework is Pytorch $2.0.0$. Note that the training phase of the diffusion model is done at the cloud server of the BS, and the receiver is only involved in the reverse inference process. Compared to the pre-training phase, semantic information generation requires lower computational resources. Therefore, the receiver is sufficient to run these modules with acceptable computational latency.

We compare our proposal with classical separation-based source and semantic communication. The traditional communication model considers the well-established image codec JPEG, an image compression algorithm used in various applications such as Internet content delivery, digital photography, and medical imaging. Many fields include Internet content delivery, digital photography, and medical imaging. The channel noise or fading is then processed using a Low-Density Parity-Check code (LDPC) and Quadrature Amplitude Modulation (QAM) scheme, denoted as JPEG+LDPC+QAM. The modulation order is set to $4$. The semantic communication model employs the classical algorithm DeepJSCC. All methods consistently use the CIFAR-$10$ dataset and are selected to be trained with SNRs between $1$ dB and $13$ dB.

\subsection{Performance comparison}

\begin{figure*}[!t]
	\centering
	\subfloat[]{\includegraphics[width=1.7in]{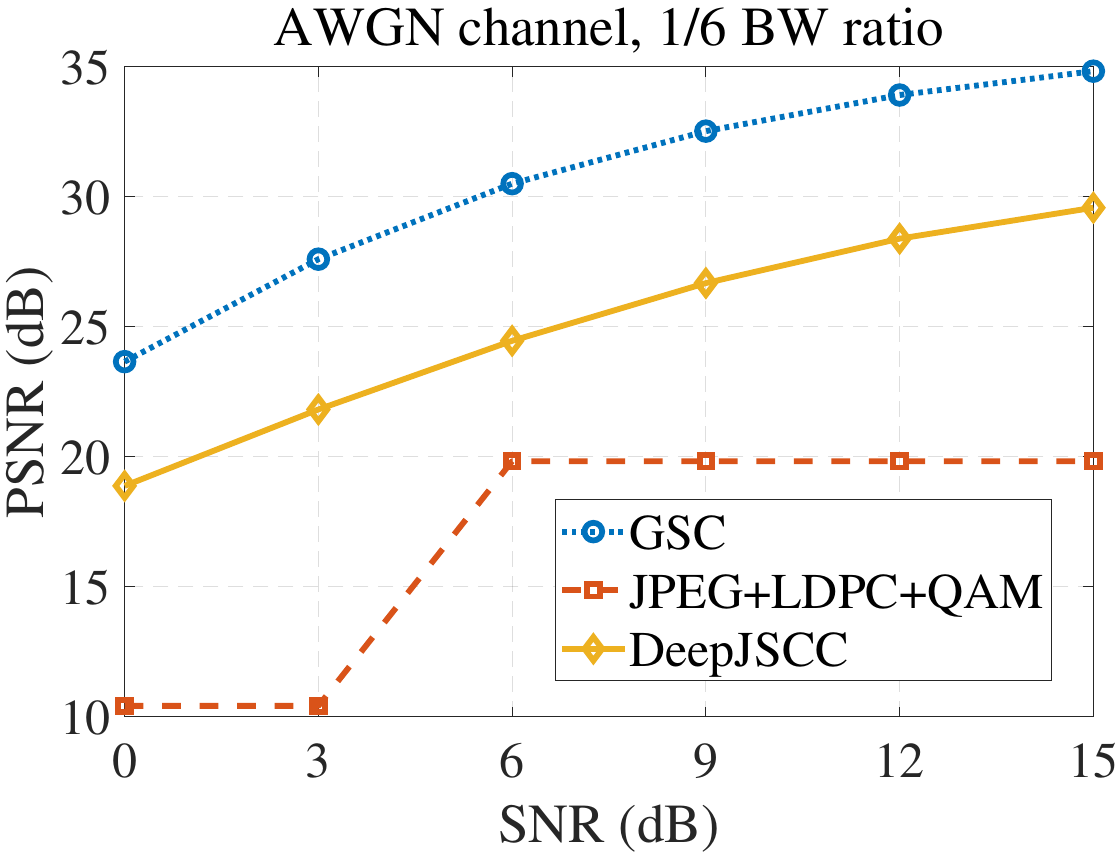}%
		\label{fig_first_case}}
	\hfil
	\subfloat[]{\includegraphics[width=1.7in]{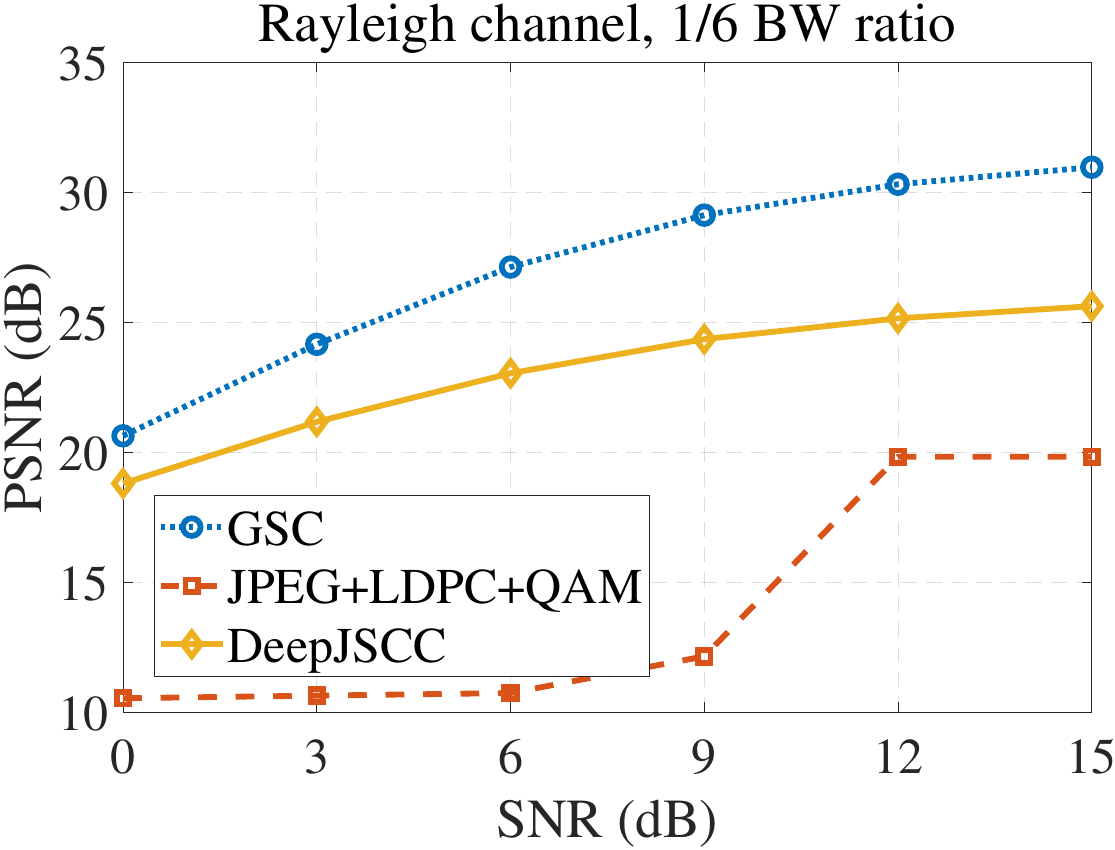}%
		\label{fig_second_case}}
	\hfil
	\subfloat[]{\includegraphics[width=1.7in]{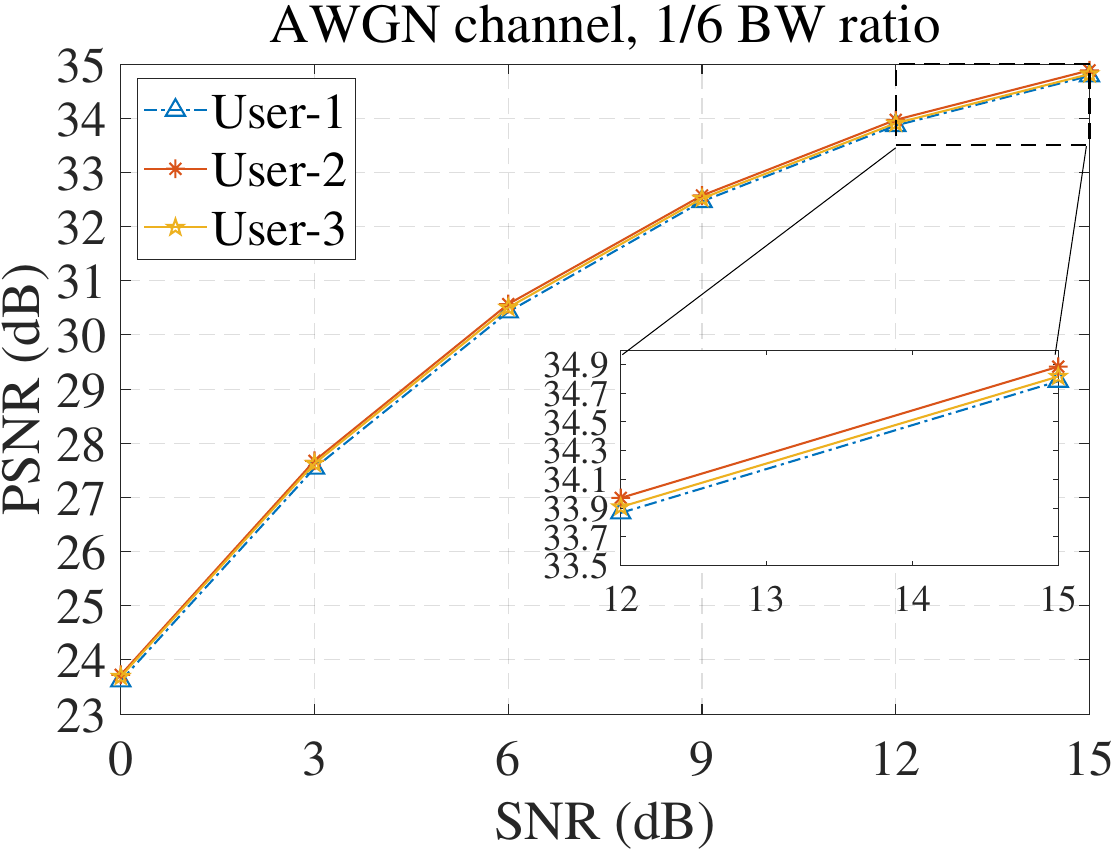}%
		\label{fig_third_case}}
	\hfil
	\subfloat[]{\includegraphics[width=1.7in]{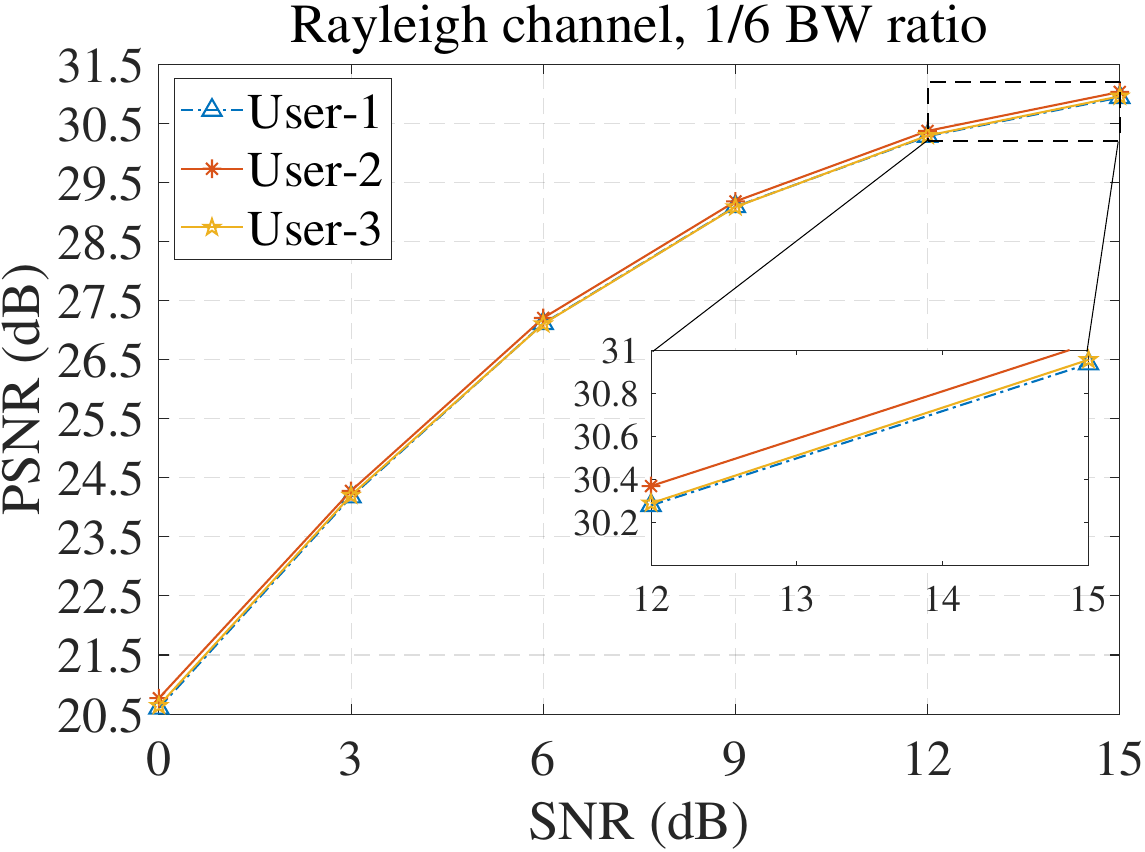}%
		\label{fig_fourth_case}}
	\caption{PSNR (higher the better) versus the SNR over AWGN and Rayleigh channel, respectively. (a) AWGN channel. (b) Rayleigh channel. (c) MU-GSC AWGN channel. (d) MU-GSC Rayleigh channel.}
	\label{fig3}
\end{figure*}

\begin{figure*}[!t]
	\centering
	\subfloat[]{\includegraphics[width=1.7in]{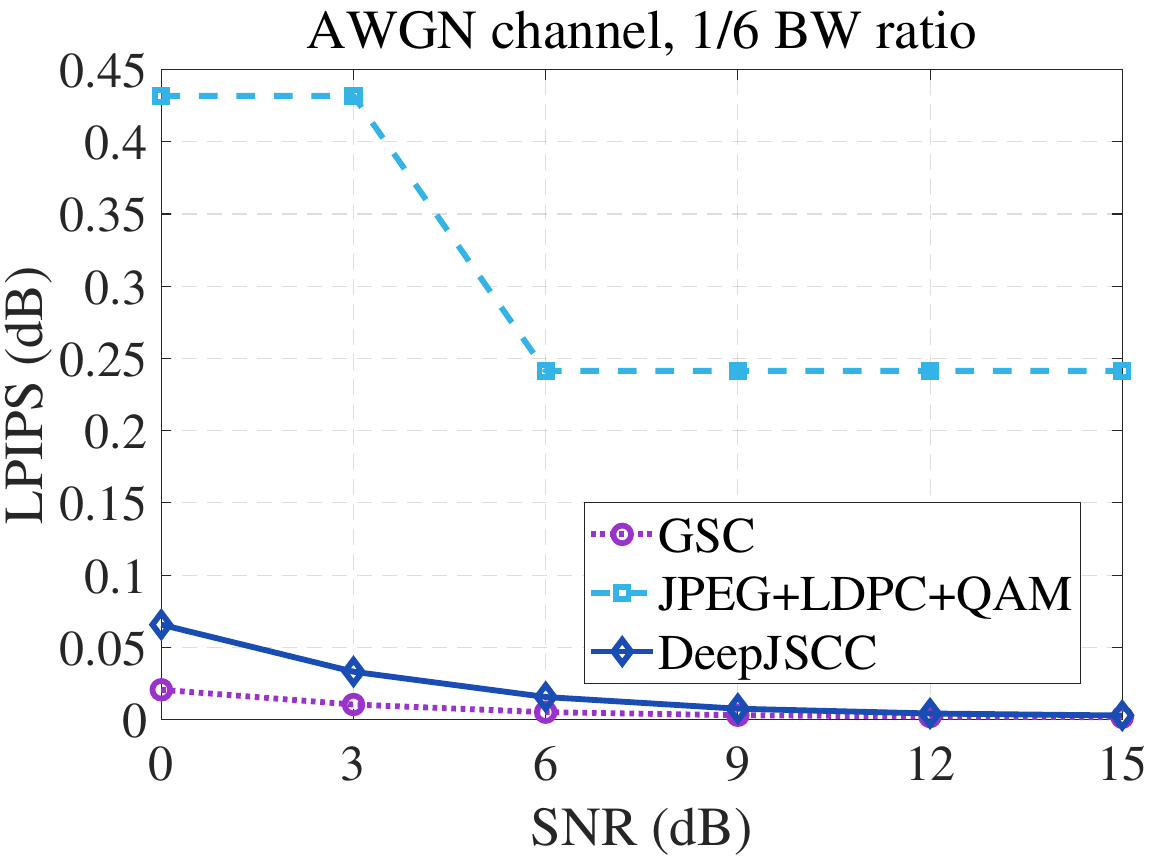}%
		\label{fa}}
	\hfil
	\subfloat[]{\includegraphics[width=1.7in]{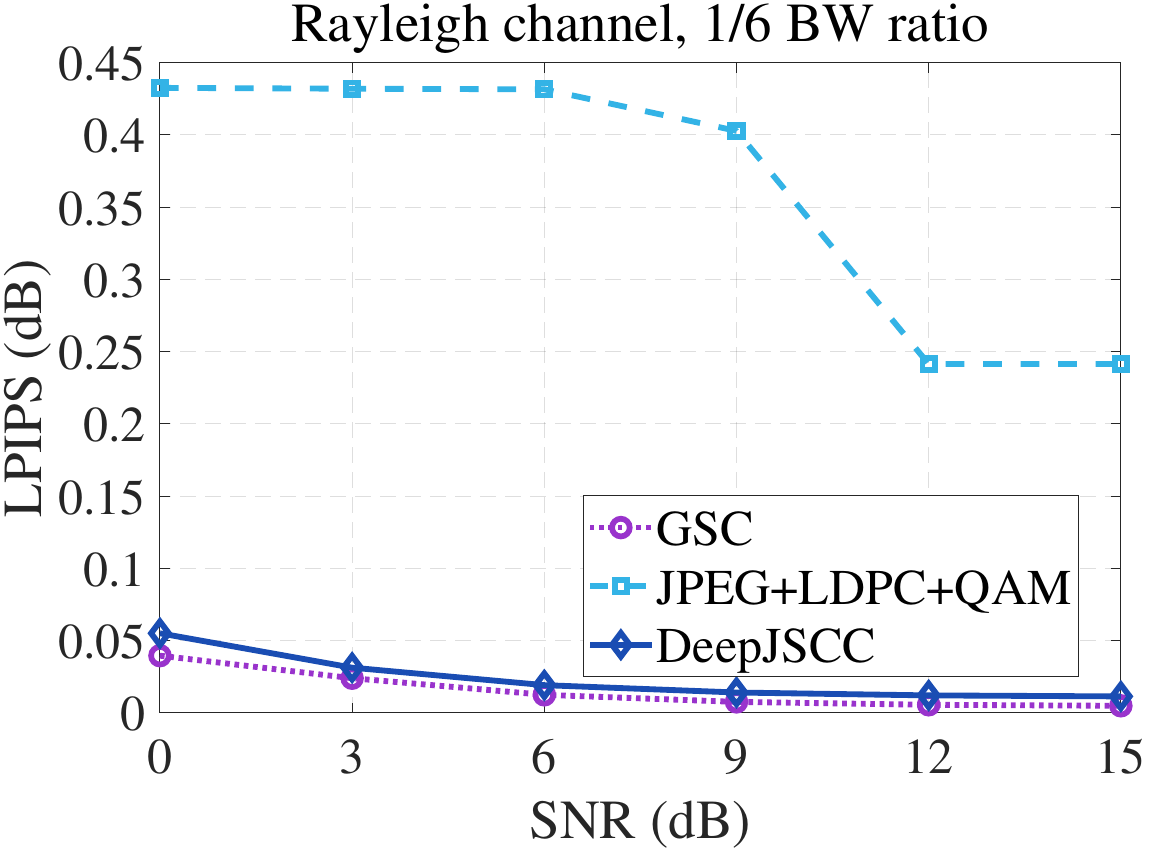}%
		\label{fb}}
	\hfil
	\subfloat[]{\includegraphics[width=1.7in]{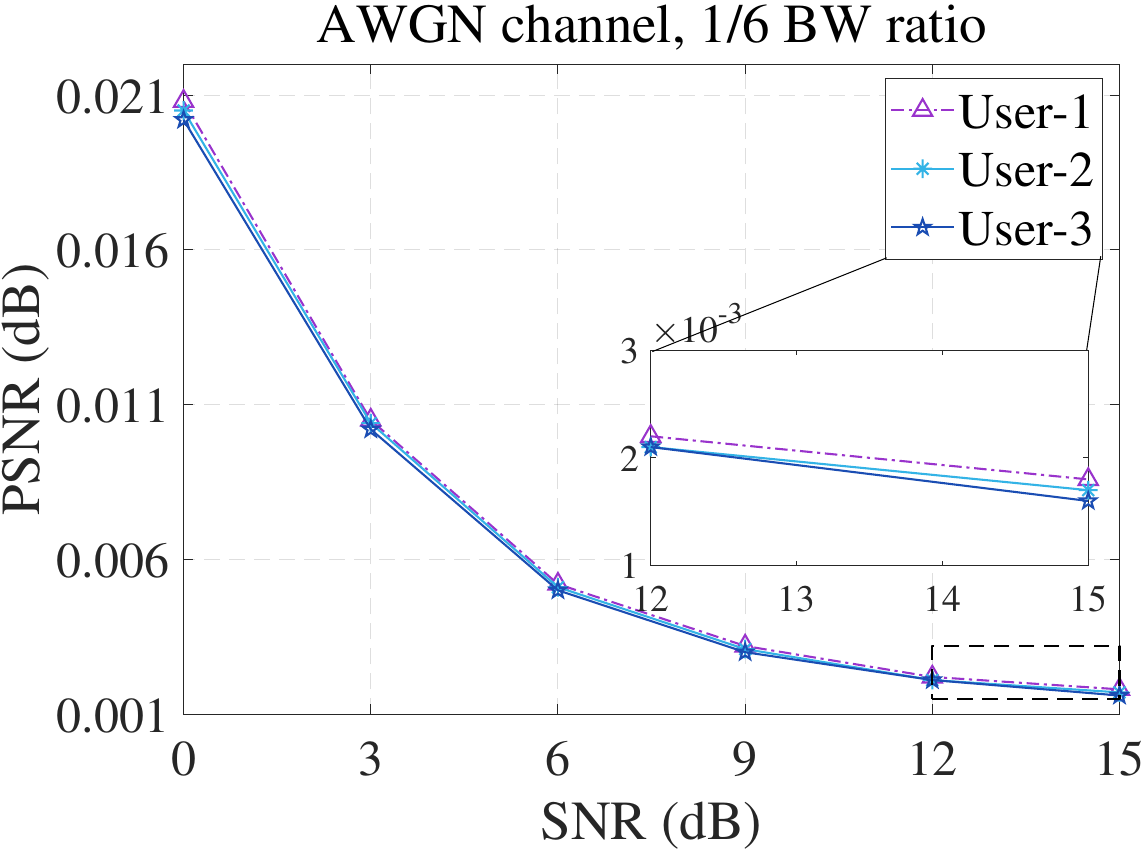}%
		\label{fc}}
	\hfil
	\subfloat[]{\includegraphics[width=1.7in]{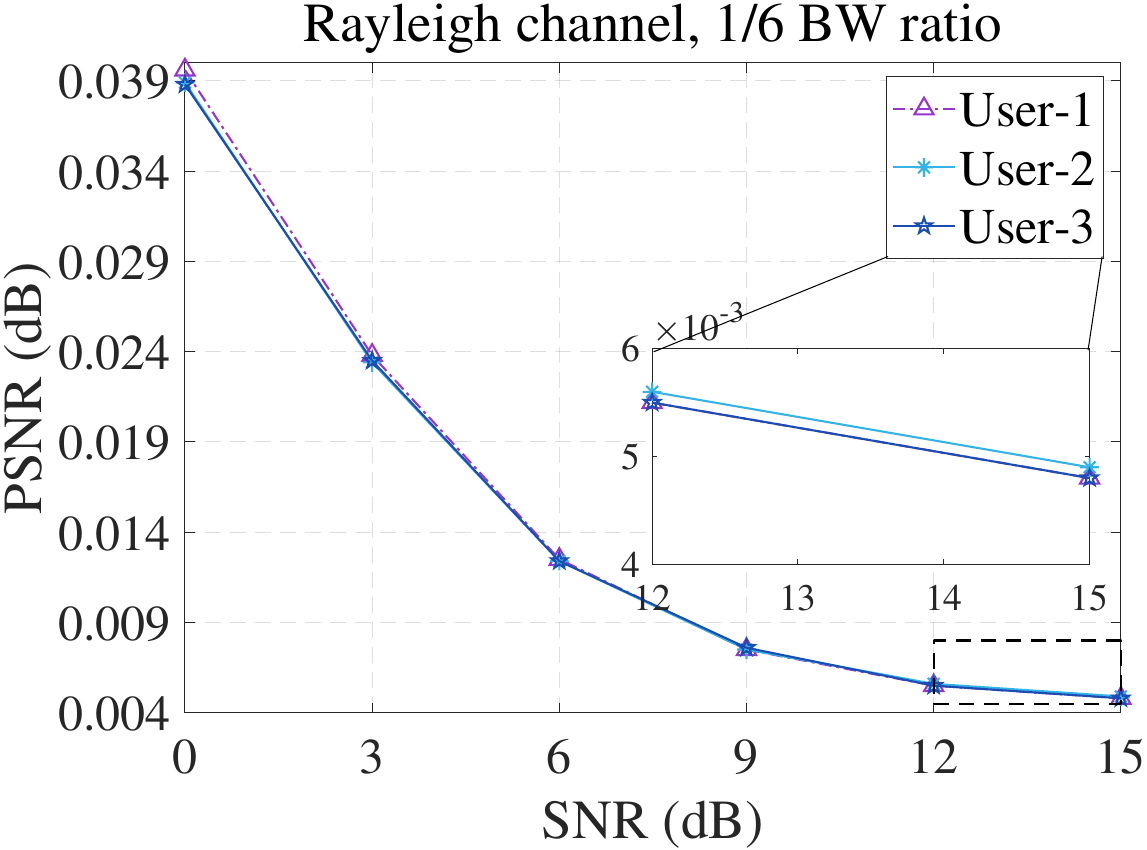}%
		\label{fd}}
	\caption{LPIPS (lower the better) versus the SNR over AWGN and Rayleigh channel, respectively. (a) AWGN channel. (b) Rayleigh channel. (c) MU-GSC AWGN channel. (d) MU-GSC Rayleigh channel.}
	\label{fig4}
\end{figure*}

To demonstrate the effectiveness of the proposed method, this section analyzes the quality of the images generated by the three methods under two different channel conditions. Specifically, we present the Peak Signal-to-Noise Ratio (PSNR) of the decoded image obtained using these approaches.

For the proposed method, a single model covers a range of SNR from $0$ dB to $15$ dB. As expected, the proposed algorithms agree with the trend of PSNR variation in semantic communication with increased SNR. Fig. \ref{fig_first_case} shows the PSNR score versus the SNR achieved by using the proposed GSC and the benchmark systems under the AWGN channels. Note that for the traditional method, i.e., JPEG+LDPC+QAM, When the channel deteriorates beyond a threshold (SNR\(<\)3), the receiver cannot decode the channel code and, therefore, cannot transmit any semantic information. Comparatively, when the SNR\(>\)6, the PSNR reaches the performance saturation of traditional communication algorithms, and the image similarity score of these methods in Fig. \ref{fig3} almost converges to 20, and further enhancement will not improve the output quality. However, with the reduction of SNR, the performance of the traditional methods is significantly degraded and is obviously poorer than that of the semantic communication systems. Our proposed system shows the same behavior as the DeepJSCC method. However, since the semantic information can be enhanced by the fine-tuning module, it is clear that the proposed GSC is more competitive than the DeepJSCC in the low SNR regime. Taking SNR=$15$ as an example, our method achieves a significant improvement in PSNR compared to the benchmark DeepJSCC algorithm, with 17.75\% of the enhancement observed in AWGN channel and 20.86\% in Rayleigh channels.

Fig. \ref{fig_second_case} shows the PSNR score versus the SNR achieved using GSC and the benchmark systems under Rayleigh fading channels. Fig. \ref{fig_second_case} exhibits the same behavior as that of Fig. \ref{fig_first_case}. Despite the more demanding harsh Rayleigh channel conditions, the GSC shows advantages in semantic communication. This observation can be attributed to the fact that, although LDPC codes and QAM modulation enhance the robustness of data transmission, JPEG compression algorithms are lossy and may lead to irreversible information loss, which reduces the fault-tolerance of the whole system, resulting in degradation of image quality or data integrity in the presence of transmission errors. Meanwhile, the PSNR of JPEG+LDPC+QAM remains constant when the channel conditions are very bad or substantially improved, which reflects the so-called cliff effect. In addition, in the forward channel, SNR ranges from 0 dB to 6 dB. Compared to the transmission performance of JPEG+LDPC+QAM, the GSC does not degrade rapidly, demonstrating a significant graceful degradation advantage. Compared with DeepJSCC, the proposed method has a lower performance gap at lower SNR, indicating that even if the received semantic information image is severely corrupted, it can still generate a realistic image consistent with the original transmitted semantic information. It is clear that our proposed system is more competitive and robust in poor channel environments. 

Figs. \ref{fig_third_case} and \ref{fig_fourth_case} show the PSNR score versus the SNR over the AWGN and Rayleigh channel, respectively, by using the proposed MU-GSC. Using the communication process of three users as an example, it is observed that the PSNR score trends similarly to the GSC. This is because the dataset and simulation setup for the multi-user system are the same as those used for the single-user system and have the same performance advantages in semantic extraction and compression.
Our proposed multi-user system outperformed in all regimes and is close to 35 when SNR = 15 dB. Meanwhile, the PSNR score of the MU-GSC means that the semantic information contained in messages is transmitted successfully. 

\begin{figure*}[!t]
	\centering
	\subfloat[]{\includegraphics[width=1.16in]{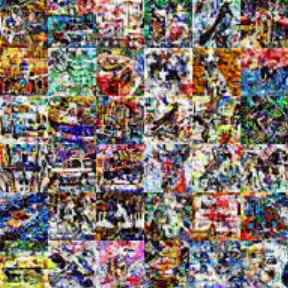}%
		\label{f1}}
	\hfil
	\subfloat[]{\includegraphics[width=1.16in]{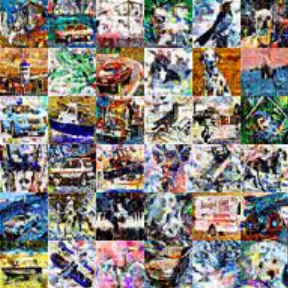}%
		\label{f2}}
	\hfil
	\subfloat[]{\includegraphics[width=1.16in]{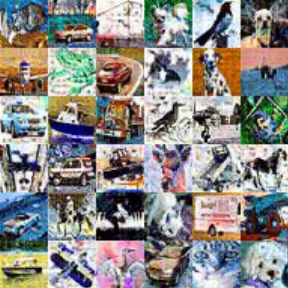}%
		\label{f3}}
	\hfil
	\subfloat[]{\includegraphics[width=1.16in]{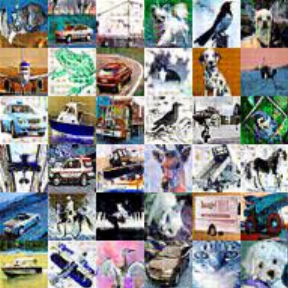}%
		\label{f4}}
	\hfil
	\subfloat[]{\includegraphics[width=1.16in]{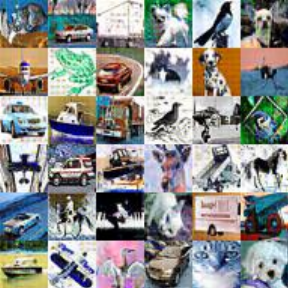}%
		\label{f5}}
	\hfil
	\subfloat[]{\includegraphics[width=1.16in]{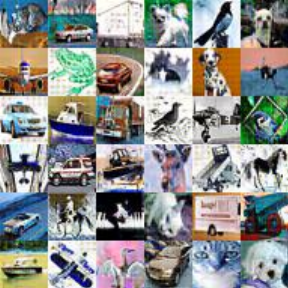}%
		\label{f6}}
	\hfil
	\subfloat[]{\includegraphics[width=1.16in]{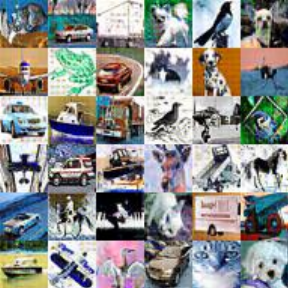}%
		\label{f11}}
	\hfil
	\subfloat[]{\includegraphics[width=1.16in]{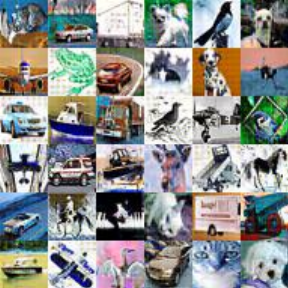}%
		\label{f22}}
	\hfil
	\subfloat[]{\includegraphics[width=1.16in]{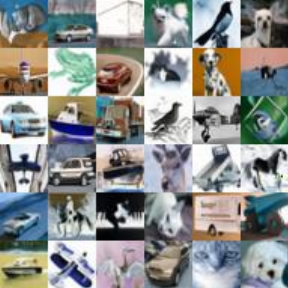}%
		\label{f33}}
	\hfil
	\subfloat[]{\includegraphics[width=1.16in]{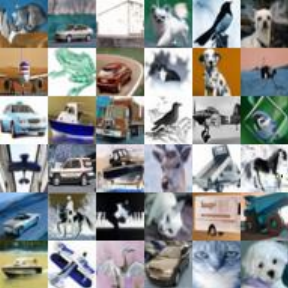}%
		\label{f44}}
	\hfil
	\subfloat[]{\includegraphics[width=1.16in]{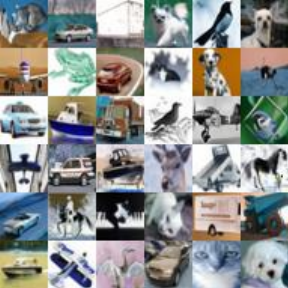}%
		\label{f55}}
	\hfil
	\subfloat[]{\includegraphics[width=1.16in]{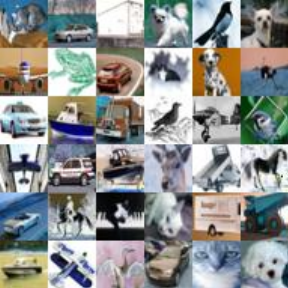}%
		\label{f66}}
	\caption{Image detail delivery of two transmission frameworks under different SNRs in AWGN channel. (a-f) NGF. (b-l) GSC.}
	\label{fig5}
\end{figure*}

\begin{table}[!t]
	\centering
	\caption{Runtime comparison of various deep learning-based systems}
	\label{tab:runtime_comparison}
	\begin{tabular}{c c c c}
		\toprule
		Models & GSC & MU-GSC \\ 
		\midrule
		Runtime (s) & 15.66 & 8.87 \\ 
		\bottomrule
	\end{tabular}
\end{table}

To more accurately assess the visual similarity of the delivered images, Fig. \ref{fig4} compares the Learned Perceptual Image Patch Similarity (LPIPS) scores versus the SNR achieved by using the proposed method and the benchmark systems under different channel conditions. Unlike the PSNR metric, which primarily measures pixel-level differences, LPIPS focuses on visual perceptual differences, where a lower LPIPS value indicates better visual perceptual quality. Figs. \ref{fa}-\ref{fb} show that the proposed GSC method outperforms existing methods in terms of LPIPS. Figs. \ref{fc}-\ref{fd} indicate that the extracted and transmitted semantic information from MU-GSC is highly compact, preserves perceptual content, and proves that the proposed method is highly robust to noise interference. Specifically, compared to the benchmark algorithm DeepJSCC, in the low SNR regime (SNR = 0 dB), the LPIPS score decreased by 68.65\% in the AWGN channel and by 28.18\% in the Rayleigh channel.

To verify the role of the semantic fine-tuning module, we performed ablation tests to corroborate the proposed method, comparing the image transmission performance of the proposed GSC method with Non-Generative Framework (NGF). 
The NGF provides a direct SemCom receiver, similar to GSC, but with the semantic fine-tuning module turned off.
Taking the example in AWGN channel transmission, Fig. \ref{fig5} allows for a visual inspection of the details of the delivered images produced by the proposed method GSC and NGF at different SNRs of the wireless channel. 
It is evident that as the SNR improves (From left to right, the SNR ranges from 0 dB to 15 dB), the images from both frameworks transition from a mottled, mosaic-like appearance to a smoother texture. 
Additionally, the structure of the GSC images becomes nearly identical to the source image. 
However, images produced by NGF display prominent issues such as spot noise and edge blurring. 
These issues arise because the semantic fine-tuning module effectively reduces channel noise and preserves valuable semantic information. 
Meanwhile, the GSC relies heavily on the image communication network, which operates under strict constraints to ensure the reliability of the transmitted images.
The results further demonstrate that the proposed algorithm is effective and robust in generating high-quality images.

The computational complexities of the proposed GSC and MU-GSC are compared in Table~\ref{tab:runtime_comparison} in terms of the average processing runtime per image. This simulation is performed with Intel Core i9-13900HX@2.20 GHz and NVIDIA GeForce RTX 4060. Compared to GSC, MU-GSC completes processing multiple user requests in nearly half the time, significantly enhancing system response speed and throughput. This improvement is attributed to the application of critical technologies such as asynchronous task processing, parallel task execution, and caching mechanisms, effectively reducing the system's computational load. However, the runtime of the proposed method is high, the reason is that the computational complexities of our proposed semantic fine-tuning algorithm increase with the size of the users' knowledge base, and this marginal cost is offset by the significant performance improvements.

\section{Conclusion}\label{sec5}
This paper proposes a generative AI semantic communication system that introduce an advanced and interpretable semantic fine-tuning module to enhance semantic information. Simulation results demonstrate that our method delivers superior transmission quality compared to traditional separation-based method and DeepJSCC, significantly improving communication services in resource-limited wireless networks, particularly under low signal-to-noise ratio conditions. Moreover, Although our method's running time is slightly higher than that of DeepJSCC, fundamental techniques such as asynchronous processing substantially enhance the system's response speed and throughput, demonstrating the scalability of the proposed single-user system in multi-user scenarios.

\newpage

\vfill

\end{document}